\documentclass[letterpaper, 10 pt, conference]{ieeeconf}

\IEEEoverridecommandlockouts

\usepackage{graphics} 
\usepackage{epsfig} 
\usepackage{mathptmx} 
\usepackage{times} 
\usepackage{amsmath} 
\usepackage{amssymb}  
\usepackage{color}
\usepackage{algpseudocode}
\usepackage{algorithm}
\usepackage{balance}

\definecolor{dkgreen}{rgb}{0,0.6,0}
\definecolor{gray}{rgb}{0.5,0.5,0.5}
\definecolor{mauve}{rgb}{0.58,0,0.82}
\usepackage{ifthen,version}
\newboolean{include-notes}
\setboolean{include-notes}{true}
\newcommand{\jknote}[1]{\ifthenelse{\boolean{include-notes}}%
{\textcolor{mauve}{\emph{JK says:#1}}}{}}
\newcommand{\tmnote}[1]{\ifthenelse{\boolean{include-notes}}%
{\textcolor{dkgreen}{\emph{TDM says:#1}}}{}}
\newcommand{\rtnote}[1]{\ifthenelse{\boolean{include-notes}}%
{\textcolor{red}{RLT says:#1}}{}}
\newcommand{\msnote}[1]{\ifthenelse{\boolean{include-notes}}%
{\textcolor{blue}{\emph{MS says:#1}}}{}}

\newtheorem{definition}{Definition}

\title{\LARGE \bf
Automated Gait Generation For Walking, Soft Robotic Quadrupeds
}

\author{Jake Ketchum$^{1}$, Sophia Schiffer$^{1}$, Muchen Sun$^{1}$, Pranav Kaarthik$^{1}$, Ryan L. Truby$^{1,2}$, Todd D. Murphey$^{1}$
\thanks{$^{1}$Center for Robotics and Biosystems, Northwestern University, Evanston, IL, USA.}%
\thanks{$^{2}$Department of Materials Science and Engineering, Northwestern University, Evanston, IL, USA.}
}

\begin{document}

\maketitle
\thispagestyle{empty}
\pagestyle{empty}

\begin{abstract}
Gait generation for soft robots is challenging due to the nonlinear dynamics and high dimensional input spaces of soft actuators. Limitations in soft robotic control and perception force researchers to hand-craft open loop controllers for gait sequences, which is a non-trivial process. Moreover, short soft actuator lifespans and natural variations in actuator behavior limit machine learning techniques to settings that can be learned on the same time scales as robot deployment. Lastly, simulation is not always possible, due to heterogeneity and nonlinearity in soft robotic materials and their dynamics change due to wear. We present a sample-efficient, simulation free, method for self-generating soft robot gaits, using very minimal computation. This technique is demonstrated on a motorized soft robotic quadruped that walks using four legs constructed from 16 ``handed shearing auxetic" (HSA) actuators. To manage the dimension of the search space, gaits are composed of two sequential sets of leg motions selected from 7 possible primitives. Pairs of primitives are executed on one leg at a time; we then select the best-performing pair to execute while moving on to subsequent legs. This method---which uses no simulation, sophisticated computation, or user input---consistently generates good translation and rotation gaits in as low as 4 minutes of hardware experimentation, outperforming hand-crafted gaits. This is the first demonstration of completely autonomous gait generation in a soft robot. 

\end{abstract}

\section{Introduction}

Soft robots---which incorporate soft actuators and compliant elements into their bodies---are a frontier for the robotics field. Flexible morphologies enable soft robots to passively adapt to unstructured environments, resist particulate damage, and reversibly recover from large or unanticipated external loads which would critically damage comparable rigid-body robots~\cite{tolley_resilient_2014}. Contact compliance also makes soft robots appropriate for interacting with delicate objects or environments. In the context of a soft quadruped, redundantly actuated soft legs can allow a robot to degrade gracefully, continuing to operate at reduced capacity after damage. As a result, terrestrially locomoting soft robots demonstrate potential in high-consequence applications ranging from inspection (where fitting into irregular geometries is common) and repair (where compliance matters) to exploration in search and rescue settings featuring severely degraded environments.

\begin{figure}[h]
    \centering
    \includegraphics[width=.45\textwidth]{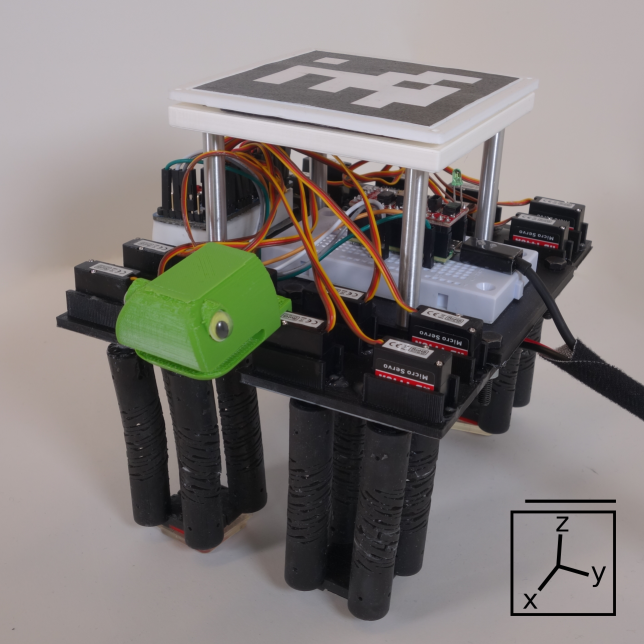}
    \caption{The HSABot used for experiments. Each leg is composed of four handed shearing auxetic actuators, each of which are controlled by micro-servos. The axes in the lower right hand corner correspond to the robot's tracking frame, which is located in the top-center of the AprilTag. The HSABot is 15cm long (x) by 13cm wide (y). The scale bar is 3cm.}
    \label{fig: robot_portrait}
\end{figure}

The key to unlocking this potential is the ability to control the motion of soft robots. Unfortunately, legged soft robots pose unique challenges when it comes to discovering propulsion strategies. Non-linear actuator dynamics and high-dimensional control spaces make hand-crafting gaits challenging. In practice, many soft robots are either only designed for forward motion, or are only equipped with forward gaits, sharply limiting their utility~\cite{goswami_3darchitected_2019, kaarthik_motorized_2022, baines_multi-environment_2022, shepherd_multigait_2011, matia_harnessing_2023, he_electrically_2019}. When accurate simulations can be built, machine learning is a valuable tool for discovering soft robotic gaits. However, when an actuator's dynamics vary substantially during a robot's life---or between different copies of the same robot---it may simply not be feasible to build and update accurate simulations. If gathering sufficient data for learning is infeasible, rigid-body approximations of the soft robot can be used to learn instead. Then sim-to-real techniques in machine learning could in principle be applied, though this limits discovered gaits to rigid body derivatives. When skilled practitioners design navigation strategies by hand, they do so using their own experience with soft materials, as well as human intuition about how compliant organisms locomote in nature. Autonomous gait-search methods must accomplish the same task without the benefit of that prior knowledge, and in a severely data-constrained environment. 

\begin{figure*}[hbt!]
    \centering
    \includegraphics[angle=0, width=1\textwidth]{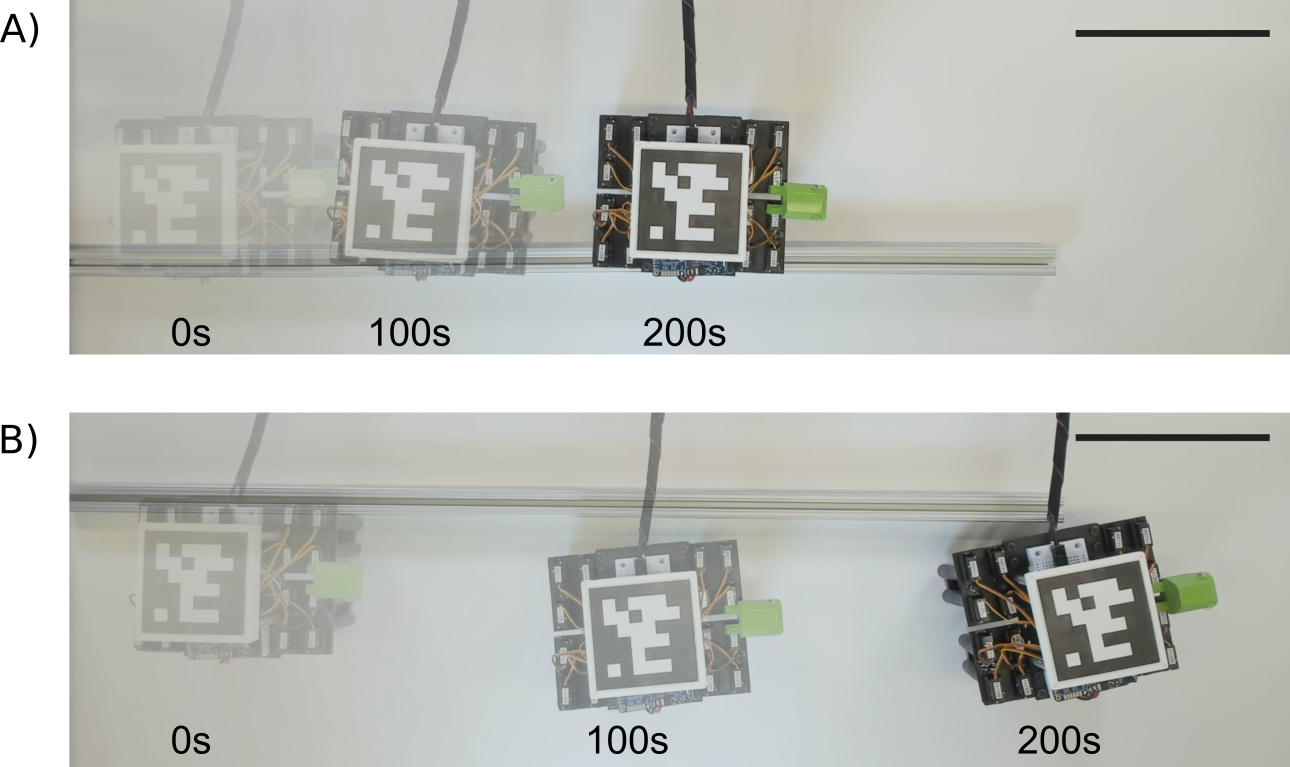}
    \caption{A ``race'' between the HSABot (A) using the hand-crafted gait from \cite{kaarthik_motorized_2022} and (B) our generated forward gait. In unconstrained space, the data-driven gait is 2.5 times faster than its hand-crafted counterpart with the same step frequency. With the aluminum guide bar shown, the data-driven gait is $\approx$2 times faster. This figure is a composite of tracking images from 0s, 100s and 200s into each run. The scale bar is 15cm---one body length---long. A guide-rail was used in both experiments.}
    \label{fig: race}
\end{figure*}

We present a straightforward, time-efficient strategy for enabling soft robotic quadrupeds to learn how to walk, with short learning time horizons and dramatic improvements in performance. Our main contribution is a tree-search based workflow for finding new gaits on-board in as little as 4 minutes. We use this method to produce a small-time locally controllable gait set for a soft quadrupedal robot, and then demonstrate a letter tracing task under closed loop control. These tasks are carried out on the HSABot, a soft quadrupedal robot which uses handed sheering auxetic (HSA) actuators to provide motive force (see Fig. \ref{fig: robot_portrait}) \cite{kaarthik_motorized_2022}. We provide a comparison to the fastest existing (hand-crafted) gait for the HSABot \cite{kaarthik_motorized_2022} and show that our method can more than double the robot's top speed without changing the step delay.

This paper is organized as follows: Section \ref{section: Related Work} discusses gait design, soft quadrupeds, and the soft actuators used in this paper. Section \ref{section: System Overview} details the robot design and experimental setup used. Sections \ref{section: Algorithm} and \ref{section: Closed Loop Control} then cover our gait search algorithm and closed loop scheduling controller. Finally, Section \ref{section: Experimental Results} discusses our experiments with generating gaits as well as racing and tracing tasks.

\section{Related Work}
\label{section: Related Work}

Legged locomotion is an active area of research for both rigid and compliant robot designs. The control of soft legged robots poses a particular challenge because key assumptions from rigid robotics, like consistent dynamics and the availability of kinematic models, do not hold. Our method rests principally on existing work in the areas of gait design, HSA actuator development, and soft robotic quadrupeds, each of which is explored in more detail below. 

\subsection{Soft Quadrupeds} 
Soft robotic quadrupeds vary in morphology and design depending on the actuation methods they employ. Early examples of soft robotic quadrupeds used fluidic elastomer actuators for complex, bioinspired walking~\cite{tolley_resilient_2014, shepherd_multigait_2011,  drotman_3d_2017, drotman_electronics-free_2021}. Others used artificial muscles based on liquid crystal elastomers, shape memory alloys, and cable tendon actuators~\cite{he_electrically_2019, tanaka_dynamic_2021, huang_highly_2019, huang_chasing_2018, ji_autonomous_2019, gong_untethered_2021, bern_trajectory_2019}. Soft robotic quadrupeds have also used thermally~\cite{baines_multi-environment_2022} or electrically~\cite{kaarthik_motorized_2022, huang_soft_2019} controlled legs for locomotion. The HSABot used in this paper uses HSA actuators to provide both the structure and motive force for its legs \cite{kaarthik_motorized_2022}.

The elastomeric materials used to manufacture soft actuators, such as polyurethanes and silicones, are highly susceptible to damage~\cite{TERRYN2021187}, leading to low fatigue and cycle limits~\cite{miron_principles_2016}. As a result, one challenge of open loop control for gait implementation on soft quadrupeds is that it depends heavily on the damage-state of a robot's actuators.  In~\cite{tolley_resilient_2014}, a functional gait relies on the resilience of the robot. Data-driven methods, such as the one highlighted in~\cite{bing2020energy} and our method, enable the possibility to work around or even take advantage of mutations in dynamics.

\subsection{Gait Design}

Although designing gaits for quadrupedal robots is not a solved problem, it has been studied extensively. The most straightforward approach involves simple repeating patterns based on quadrupedal motion observed in nature, for example trotting, striding, galloping or crawling~\cite{shepherd_multigait_2011, bledt_mit_2018, katz_mini_2019, kau_stanford_2019}. This approach works well for basic motions, and for simple robots can even be implemented without invoking a formal kinematic model of the robot~\cite{shepherd_multigait_2011}. 

A generalization of this approach involves parameterizing gaits in terms of the trajectories followed by each foot, as seen in~\cite{bledt_mit_2018, kau_stanford_2019}. This helps abstract away some of the robot design and allows tools like optimization to be brought to bear more easily. However, it does require at least a basic kinematic model of the robot.

Other more sophisticated methods have also been used. For example, contact planning---in which ground capture is used to plan footfalls along some desired trajectory---has proven successful in helping robots navigate complex environments~\cite{geisert_contact_2019}. Similarly, reinforcement learning (RL), often trained in simulation, can provide improvements in gait speed and robustness~\cite{margolis_rapid_2022}. However, these methods almost universally require good robot models to function well. 

Soft robots most often use either bioinspired gaits~\cite{shepherd_multigait_2011, matia_harnessing_2023} or some version of foot-trajectory planning. The former is common in soft robotics because it lends itself to hand-crafting. Contact planning assumes a level of autonomy which is rare in soft robotics, but RL has been used for some soft robots~\cite{ji_synthesizing_2022, ishige_learning_2018}. Unfortunately, for soft robots which require finite element methods for accurate simulation, gathering enough simulated data to use machine learning effectively can be challenging. In this paper, we take an intermediate approach, using data-driven methods to find the gait while restricting the search space to simple repeating patterns. 

\subsection{Handed Shearing Auxetic Actuators} 

Handed shearing auxetics (HSAs) are a class of architected materials that has been engineered for motorized, soft robotic actuation~\cite{lipton_handedness_2018}. Cylindrical HSAs with opposite handedness have been assembled into soft robotic actuators for multi-degree-of-freedom (multi-DOF) motion and grasping~\cite{lipton_handedness_2018,chin_simple_2019}. HSAs can be 3D printed using digital projection lithography and polyurethane-based, photopolymer resins~\cite{truby_recipe_2021}. They can be fluidically innervated or equipped with internal cameras for distributed sensing capabilities~\cite{truby_fluidic_2022, zhang_vision-based_2022}. More recently, mult-DOF HSA platforms were assembled into soft robotic legs from 3D printed, miniaturized HSA pairs~\cite{kaarthik_motorized_2022}. Miniaturization allowed four HSA legs to be actuated by a total of 16 lightweight microservo motors to achieve battery-powered, untethered walking for 65min at 2 body lengths per min with a payload capacity of over 1.5kg. Unfortunately, HSA-based actuators will inevitably demonstrate time-varying mechanical behaviors given their fabrication from viscoelastic materials \cite{truby_recipe_2021}. Like the vast majority of soft robotic actuators, higher actuation speeds (i.e., higher strain rates) and strains of HSA actuators result in higher rates of degradation of mechanical stiffness and changes of repeatable actuation over time \cite{truby_recipe_2021}.

\section{System Overview and Methods}
\label{section: System Overview}

The HSABot used in this work (see Fig. \ref{fig: robot_portrait}) is adapted from~\cite{kaarthik_motorized_2022}. It is made of four HSA legs actuated by 16 microservo motors (Power HD 1440A, Pololu). The HSAs are 3D printed from a single-cure, polyurethane resin (E-Rigid PU Black, ETEC) using a digital light processing (DLP) printer (D4K Pro, ETEC)~\cite{kaarthik_motorized_2022}. All 16 servos are glued to their associated actuator, and are driven by an Adafruit PCA9685 driver board. An onboard Teensy 4.0 acts as a USB-to-I2C bridge, allowing the robot to be controlled by way of a power/USB tether mounted to the top of the test frame. The experimental code is written in Python and interfaces with the robot using a ROS driver for the onboard Teensy. ROS is also used to manage the video streams and track the robot using the \texttt{ROS\_apriltag} package. The robot has a body length (BL) of 15cm.

Throughout our experimental trials, we used four sets of HSA legs. Although no leg was damaged enough to prevent locomotion, we did observe degradation of the HSA actuators via strain-induced weakening over time in our experiments. We mitigate actuator damage over time by limiting the servos to $80\%$ of their nominal travel ($\pm 1.25rad$) to prevent over-torquing and run the gaits with a step delay of 0.25s except when testing for speed. Even with these settings, we found that gait behavior would begin to change noticeably after 1-2 hours of continuous walking as the legs weakened and became easier to actuate. This is consistent with the time-varying mechanical behaviors of viscoelastic materials used in HSAs~\cite{kaarthik_motorized_2022, truby_recipe_2021,truby_fluidic_2022} and other soft robotic actuators.

All training was conducted on either a polished wooden surface or photo paper, and tracing experiments were run photo paper. Whenever used, the photo-paper was folded and secured at the table edges to ensure consistent gait behavior. One-inch square polyurethane pads were added to the feet for grip. Though we observed more stick-slip gait behavior on the paper compared to the wood, all gaits performed acceptably on both surfaces. 
HSABot tracking was accomplished using an Intel Realsense D435 camera with a resolution of 1280 by 720 at 15hz. Two AprilTags, placed on the robot and table, provided the robot and origin frames respectively. A test cage confined the robot to a 750mm square, and held the camera above the table. Tracking within that region was repeatable to~0.5mm, and we did not observe any significant tracking noise in the positioning data. 

\section{Algorithm}
\label{section: Algorithm}

\subsection{Notation}
We denote the 16 servo positions as a vector $c=[c_1,\dots,c_{16}]\in\mathbb{R}^{16}$, and we denote the state of the robot itself as its position $x,y$ and orientation $\theta$. 

\begin{definition}[Step]

    A step is defined as one full set of servo positions $c^n \in\mathbb{R}^{16}$. 

\end{definition}

\begin{definition}[Gait]
    A gait is defined as a sequence of steps, denoted as $g = \{c^{1}\dots{c^{n}}\}$.
\end{definition}

\begin{definition}[Primitive]

    A primitive is defined as a set of 4 servo positions which cause one leg to assume a particular configuration, denoted as $p_n = [c_1, \dots, c_4]\in\mathbb{R}^{4}$. A set of 4 primitives fully defines the robot's servo positions. 

\end{definition}

\begin{definition}[Drift]

    In this work, drift refers to any motion orthogonal to the desired velocity vector for a particular gait. This can either be inherent to the gait or due to external factors. 

\end{definition}

\begin{algorithm}
\caption{Tree Search Algorithm}\label{alg:search}
\begin{algorithmic}

\State // \textit{Define set of all primitives.}
\State $P \gets \{p_0, \dots, p_6\}$
\\
\State // \textit{Initialize gait arrays to all-neutral.}
\State $G_{current}\gets[(p_0, p_0),(p_0, p_0),(p_0, p_0),(p_0, p_0)]$
\State $G_{best} \gets G_{current}$
\\
\State // \textit{Initialize reward to negative infinity.}
\State $R_{best} \gets -\infty$
\\
\State // \textit{Search on each leg in turn.}
\For{$L \in \{0, \dots, 3\}$}
    \\
    \State // \textit{Test this primitive permutation.}
    \For{$(p_a,p_b) \in P \times P$}
        
        \State $G_{current}[L] = (p_a, p_b)$
        \State $R_{current} = Evaluate Gait(G_{current})$
        \If {$R_{current} > R_{best}$}
        
            \State $R_{best} \gets R_{current}$
            \State $G_{best} \gets G_{current}$
            
        \EndIf{}

    \EndFor{}
    \State $G_{current} \gets G_{best}$
    
\EndFor{}
\end{algorithmic}
\end{algorithm}

In this work, we define a gait as being composed of three steps of servo positions separated by equal time steps ($g = \{c^{1},c^{2},c^{3}\}$). Each servo position can be either fully left (state = 0), fully right (state = 1), or centered (state = 0.5) in the servo travel. The fully left and fully right positions correspond to extension or contraction of the HSA depending on the cardinality of the actuator, and the centered position serves as a neutral default. In the first two gait steps the servos are set to positions determined by the search algorithm, and in the final stage the servos return to neutral. This allows different gaits to be sequenced arbitrarily without introducing unpredictable behavior during transitions. The step-delay is not considered a search parameter and can be varied to increase or decrease the gait velocity.

\begin{figure}[ht]
    \centering
    \includegraphics[width=0.45\textwidth]{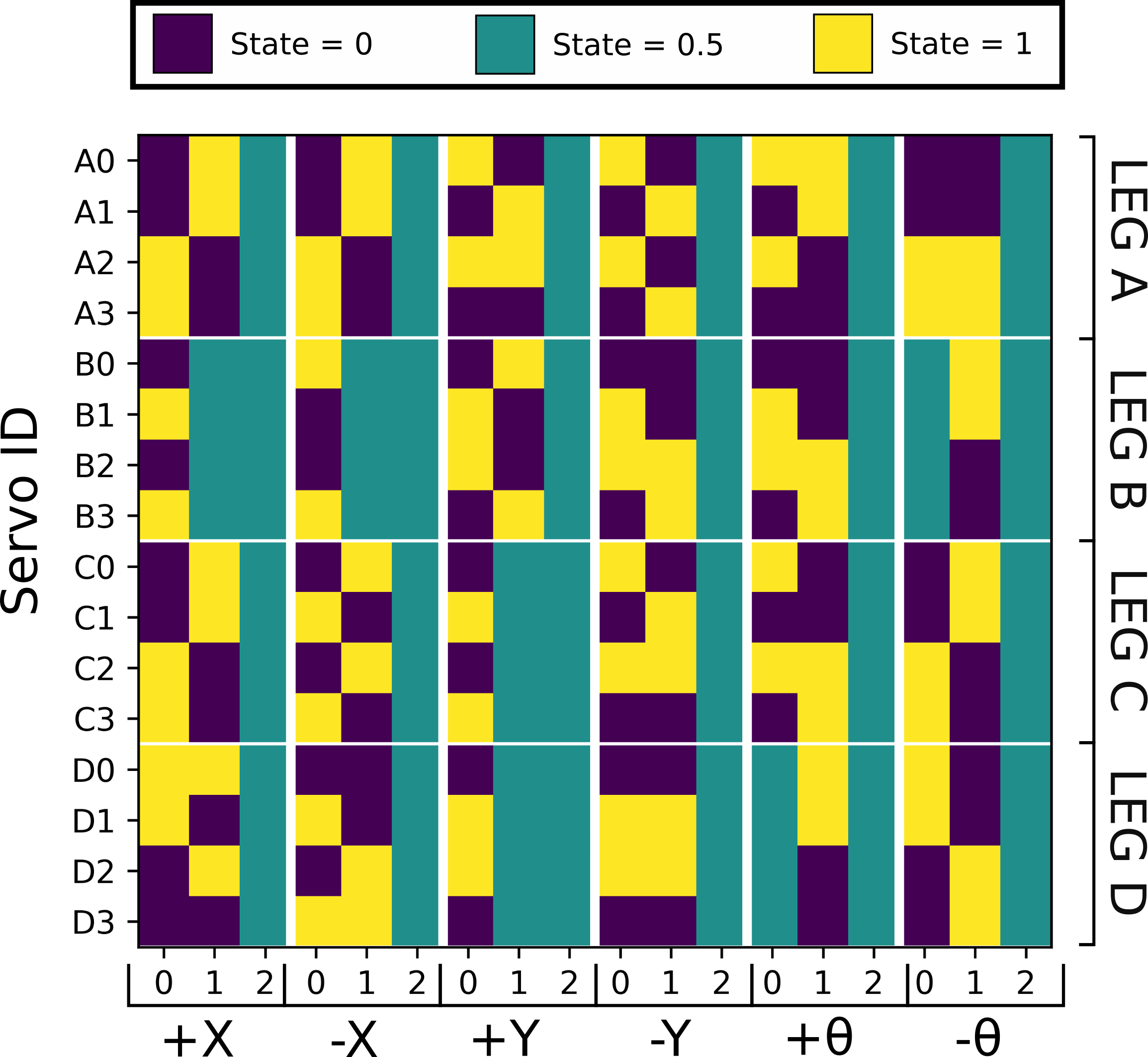}
    \caption{This figure shows 6 different gaits for translation along the robot's x and y axes, as well as in-place rotation clockwise (i.e., left turn, $+\theta$) and counter-clockwise (i.e., right turn, $-\theta$). Each gait is composed of three steps of 16 servo positions. Servo positions can be either fully left (0), centered (0.5), or fully right (1). This set of gaits corresponds to the velocities shown in Figure \ref{fig: vel_means}.}
    \label{fig:fig2}
\end{figure}

To restrict the search space for new gaits, we selected 7 leg-based primitives from among the leg positions found in~\cite{truby_recipe_2021}. A gait is thus uniquely defined by either 32 servo positions (two steps of 16), or 8 leg primitives (two steps of 4). The primitives selected are leg neutral, linear extension, linear contraction, and leg-tilt in each of the four cardinal directions. These primitives were hand-chosen to span the range of extreme motions available to each leg, although in practice, behavior varies substantially between legs, and it may reverse or become amplified as the legs wear. Since the algorithm does not assume primitives will have a particular behavior---only that the primitives induce different behaviors---these shifts are handled automatically. 

To find a new gait, we employ a depth first tree search, in which the nodes are combinations of servo positions for a particular leg. We first iterate through all 49 two-step primitive permutations for leg A and select the best. We then continue to play that sequence on leg A, as we test all 49 permutations for leg B, and so on until the entire gait has been found. This process is outlined in Algorithm \ref{alg:search}. Subsequent legs are thus searched in the context of previous legs. We do not believe that a tree search is necessarily the optimal method, but it allows us to complete a gait search in 196 evaluations, rather than the $5.8 \cdot 10^6$ required for a brute force search of all primitive combinations, or the $2^{64}$ evaluations to exhaustively search the space of all possible two-step gaits. Equation \ref{equ: N_Evals} shows the number of evaluations required, for a time complexity of $O(n)$ with respect to the number of legs and $O(n^2)$ with respect to the number of primitives:
\begin{equation}
N_{evals} = N_{Legs} N_{primatives}^2
\label{equ: N_Evals}
\end{equation}
This sample efficiency is one big advantage of our method over reinforcement learning approaches like Q learning. By limiting the exploration to less than 200 evaluations, the tree search can be used for efficient re-training in the field. 

When finding a new gait, the default servo positions are set to neutral. However, they can also be set to a previously found gait, allowing the algorithm to re-evaluate each leg's optimal behavior in the context the existing gait. This refinement process helps reduce the search-order dependence of gaits, and monotonically improves gait performance. The impact of refinement is bigger for translation heavy gaits, than for rotation heavy gaits. We recommend refining translation gaits as a matter of course. 

To evaluate each gait candidate, we use the reward function shown in equation \ref{equ: reward_function} which rewards motion along a primary axis, and penalizes drift along the other axes:
\begin{equation}
R(\dot x,\dot y, \dot{\theta}) = a \dot x + b \dot y + c \dot \theta + e |\dot x | + f |\dot y| + g |\dot \theta |
\label{equ: reward_function}
\end{equation}
Other cost functions could be used to reward different types of gaits. The velocities used are in the body-frame of the robot at the start of an evaluation.  Due to stochasticity in the robot’s motion, we have sometimes found it helpful to execute multiple gait cycles per evaluation. Using a step delay of 0.1s and 3 cycles per evaluation each gait takes just under 4 minutes to train. However, to preserve leg life we often used two rounds of training with 0.15s per step, and 2 gait cycles per evaluation. Using those parameters the process takes 7 minutes and 50 seconds to find a gait. 

\section{Closed Loop Control} 
\label{section: Closed Loop Control}

Tracing and trajectory following are key robot capabilities for completing higher-level tasks like search or mapping. This section presents a closed loop gait scheduling controller which enables the HSABot to mimic differential drive dynamics even as leg wear leads to changing gait drifts and velocities. For the purposes of this paper, a trajectory is assumed to be composed of segments which are either pure translations along the robot's forward/backward axis (x), or pure rotations about the vertical axis (z). In principle, this allows the robot---for a given error tolerance---to follow any line which is specified using (x,y) coordinates along its length. However, in practice this controller is not suitable for curve tracing.

\begin{definition}[Primary Gait]

    In any given trajectory segment the primary gait is the one which would, in the absence of drift, allow the robot to complete that segment. For example, the primary gait for a forward translation segment is $+x$ and the primary gait for a rightward turn is $-\theta$. 

\end{definition}

When the robot tracks a linear trajectory segment, it repeatedly calls either its body-fixed $+x$ (forward) or $-x$ (backward) gait until it reaches the end of the line segment. As the robot travels, its $y$ (sideways) and $\theta$ (rotational) drift is measured relative to the segment, and once the drift exceeds a specified tolerance---in our case [.05m, .05m, .05rad]---the controller schedules corrective gaits in place of the primary gait until the drift has been eliminated (e.g., if heading drift exceeds tolerance in the $+\theta$ direction, $-\theta$ gaits would be scheduled until the heading drift has been reduced to 0).  The same process is used for ensuring the robot does not drift while rotating, except that the correction gaits are $+x, -x, +y, -y$ instead of $+y, -y, +\theta, -\theta$ as they would be for linear segments.

\section{Experimental Results}
\label{section: Experimental Results}
\begin{figure}[h]
    \centering
    \includegraphics[angle=0, width=0.45\textwidth]{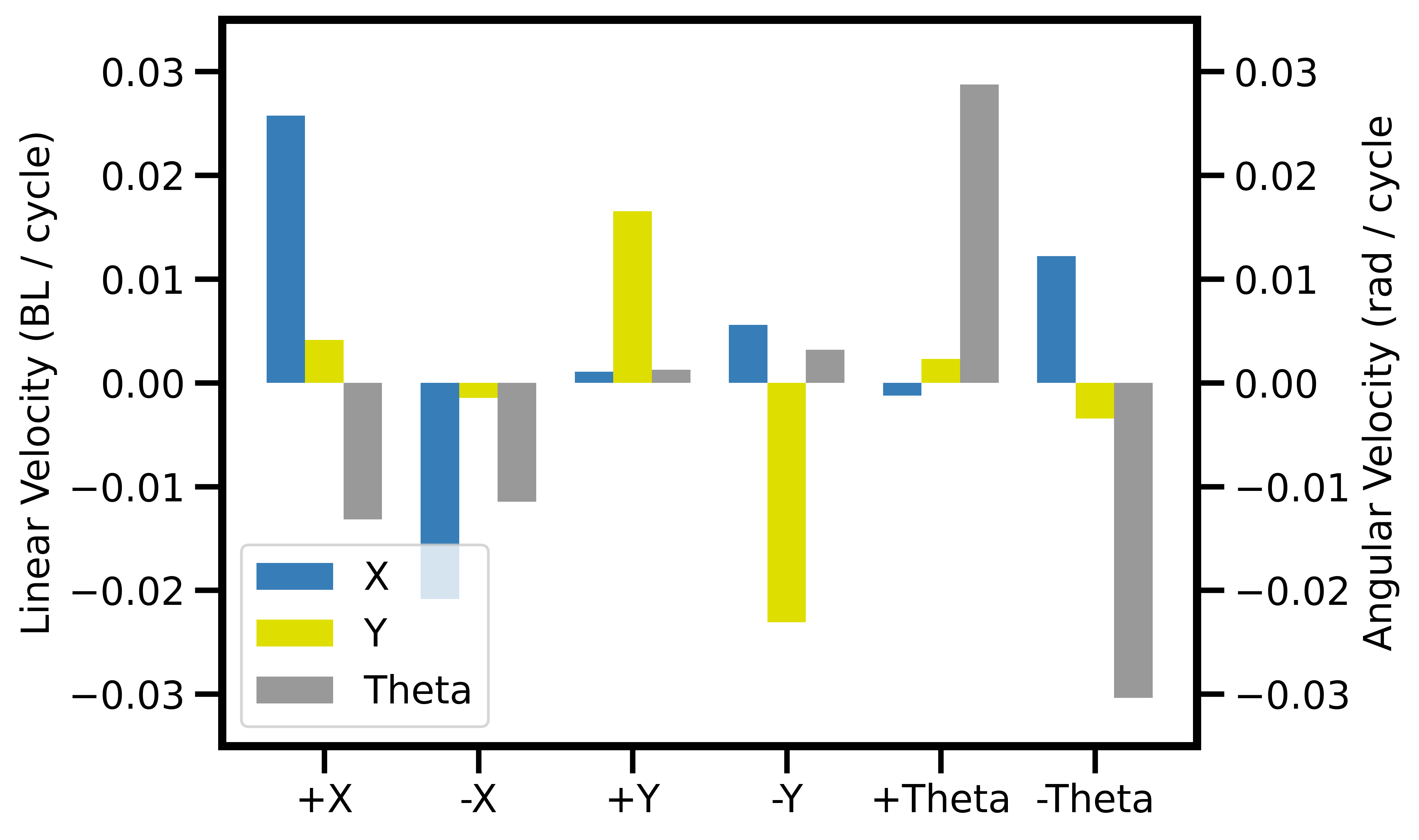}
    \caption{This chart shows the mean velocity of the robot for six body-centered axial gaits. Note the relatively high drift on the two x gaits and the $+\theta$ gait. Linear velocities are reported in body lengths (BL) per gait cycle, while angular velocities are reported in radians per gait cycle.}
    \label{fig: vel_means}
\end{figure}

\begin{figure}[h]
    \centering
    \includegraphics[angle=0, width=0.45\textwidth]{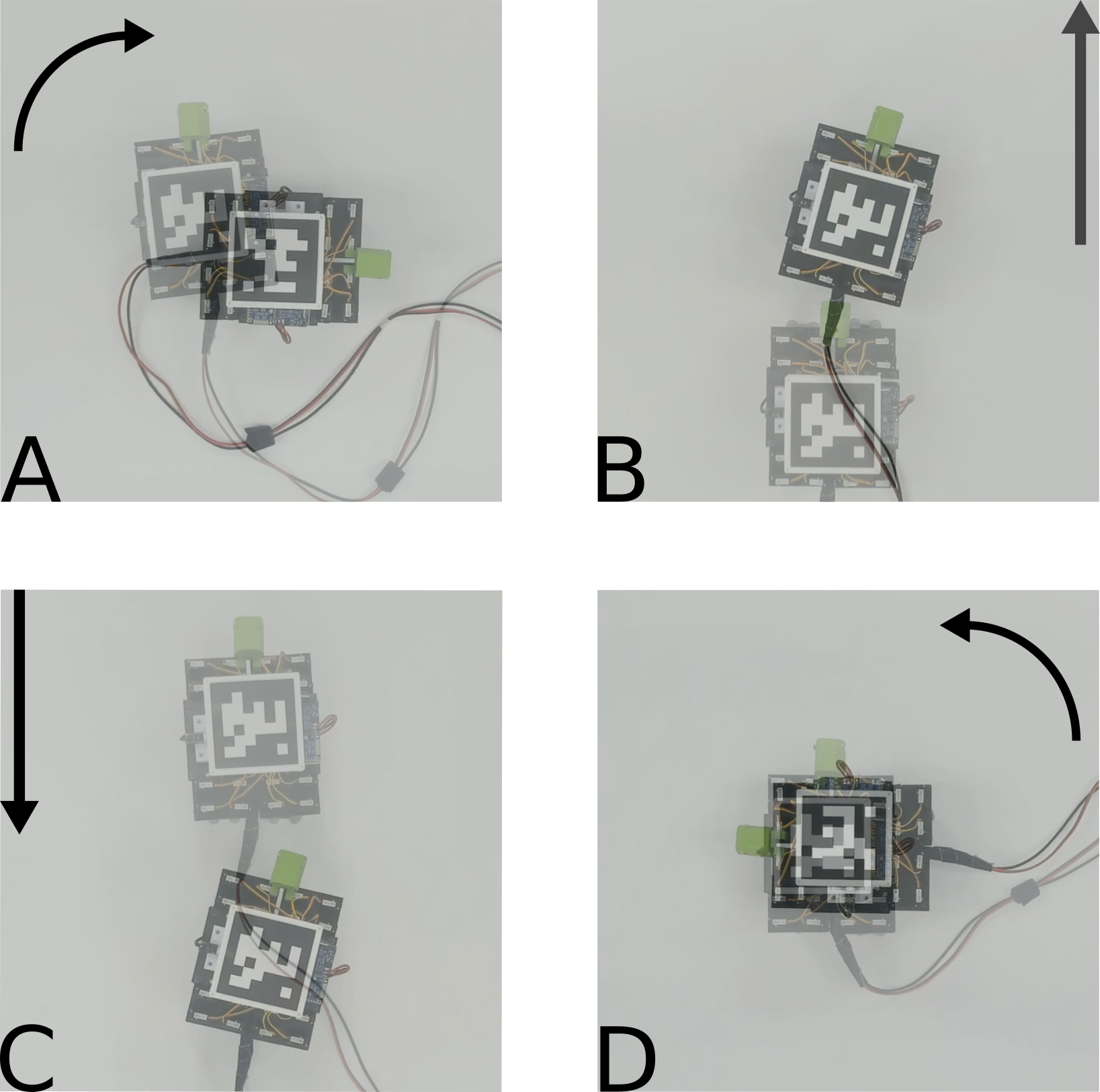}
    \caption{Four representative data driven gaits used for the N tracing figure: A) Right Turn ($-\theta$), B) Forward ($+x$), C) Backward $-x$, D) Turn Left ($+\theta$). Not shown here are the two side shuffle gaits which were also used for corrective moves during the closed loop tracking process. }
    \label{fig: 90_rotation}
\end{figure}

As shown in Figure \ref{fig: vel_means}, we trained a set of 6 gaits--Forward, Backward, Left, Right, Turn Left, and Turn Right--which allowed the HSABot to rotate and translate along its body-centered axes. The rotation gaits were trained with one cycle-per-evaluation, and the translation gaits were trained with three cycles-per-evaluation. We used the reward function coefficients found in Table \ref{tbl: reward_coefficients}, which produced fast gaits with minimal drift. Figure \ref{fig: 90_rotation} shows a 90 degree rotation performed using a data-driven $+\theta$ gait. We found that increasing coefficients ``d'', ``e'', and ``f'', helps reduce drift during refinement, but that for initial training it is best to keep them relatively low so that ``not moving at all'' is not a local maxima for the reward function. 

\begin{table}[h]
    \centering
\begin{tabular}{ c | c c c c c c }
      & a & b & c & d & e & f\\ 
 \hline
 
$+x$ (Forward) & 1 & 0 & 0 &  0 & -0.1 & -0.1 \\  
$-x$ (Backward)& -1 & 0 & 0 &  0 & -0.1 & -0.1 \\  
$+y$ (Left Shuffle)& 0 & 1 & 0 &  -0.1 & 0 & -0.1 \\  
$-y$ (Right Shuffle)& 0 & -1 & 0 &  -0.1 & 0 & -0.1 \\  
$+\theta$ (Left Turn)& 0 & 0 & 1 &  -0.1 & -0.1 & 0 \\  
$-\theta$ (Right Turn)& 0 & 0 & -1 &  -0.1 & -0.1 & 0 \\  
\end{tabular}
\caption{Reward function coefficients used for training.}
\label{tbl: reward_coefficients}
\end{table}

\begin{figure}[h]
    \centering
    \includegraphics[angle=0, width=0.45\textwidth]{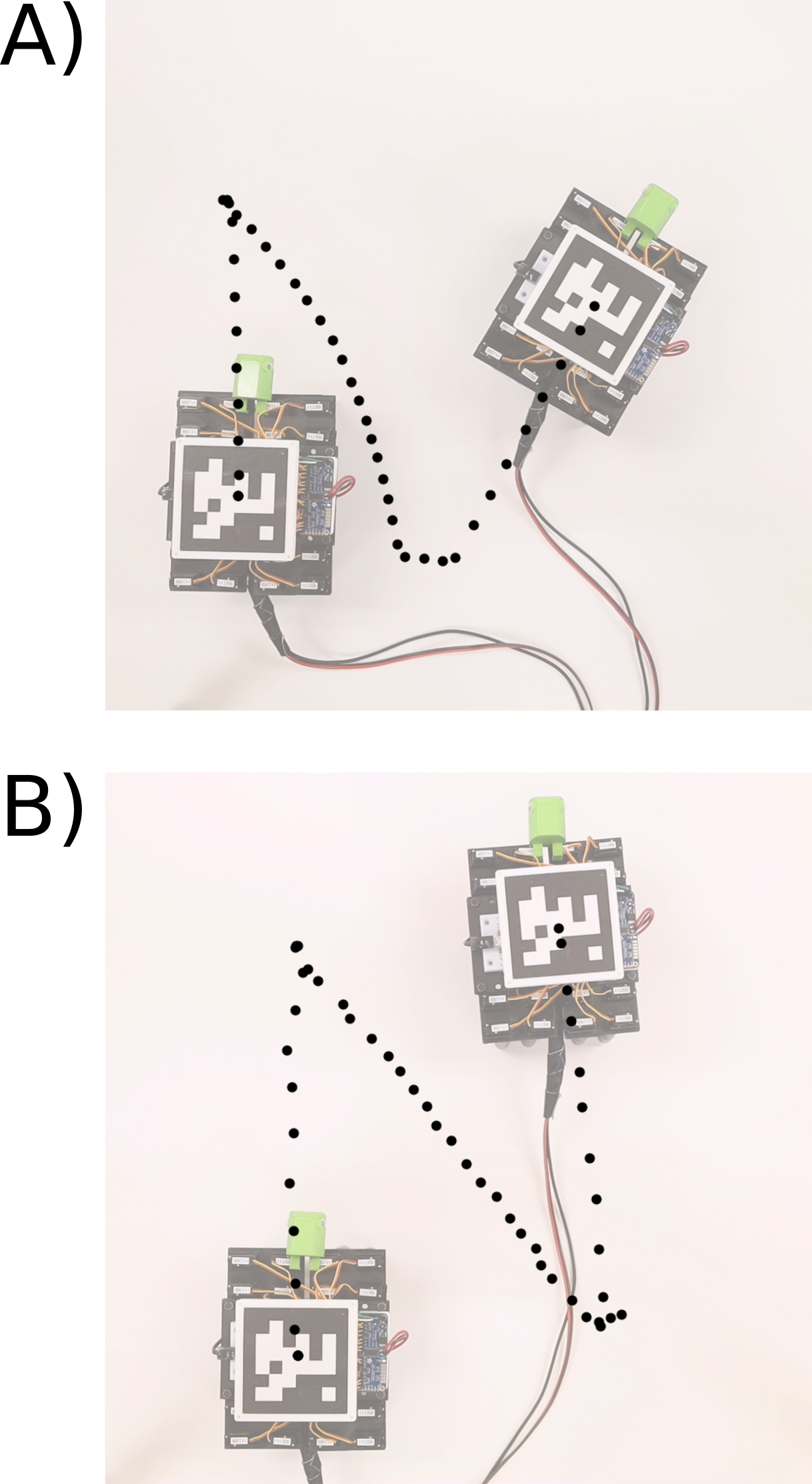}
    \caption{Using closed loop control, the robot is able to trace out non-trivial shapes. A) The HSABot traced a 20cmx15cm Capital N--- for Northwestern University ---by following an open-loop trajectory. B) The same tracing task for a 30cmx22.5cm N completed using closed-loop control. This experiment was conducted over several operating-hours after the gaits were trained and tested. By using the closed loop controller, the HSABot is able to compensate for the resulting changes in gait behavior---particularly rotation---while the open loop solution is not.}
    \label{fig: medium_n}
\end{figure}

The combination of these 6 gaits allows the robot to navigate the world and perform tracing tasks under either joystick or closed loop control. We demonstrated this by having the robot trace an N as shown in Figure \ref{fig: medium_n}. This test was run under closed loop control with similar gaits to those shown in Figure \ref{fig: vel_means} and a drift evaluation every 4 gait cycles. Figure \ref{fig: medium_n} also shows an open loop trail of the same target trajectory (an N), conducted using the mean velocities for each gait to generate a control sequence. The open-loop N was reduced in size by one-third relative to the closed-loop N --from 30cm in height to 20cm-- to ensure the robot would remain on the testing table. There was moderate post-training wear on two of the actuators, which may explain why the open loop figure (Figure \ref{fig: medium_n}A) shows gaits that are consistently faster than would be expected. 

To evaluate the speed of generated gaits, we compared a freshly generated forward $(+x)$ gait to the hand-crafted forward gait\footnote{The researchers attempted to hand-craft rotation gaits as an additional point of comparison, but were unable to do so after an order of magnitude longer than was required for the linear gaits.} from~\cite{kaarthik_motorized_2022}. Figure \ref{fig: race} shows the two gaits with snapshots of the robot every 100s. The data-driven gait was trained with one round of refinement, and we used a step size of 100ms for both gaits. The data-driven gait produced a forward velocity of 5.6mm/s (0.037 BL/s), just under 2.5 times faster than the hand-crafted gait at 2.25mm/s (0.015 BL/s). Interestingly, the hand-crafted gait was slower than the 3.4mm/s reported in~\cite{kaarthik_motorized_2022}. We think this is due to a combination of natural variability between robots and fairly substantial actuator damage on leg D. No difference in velocity was observed for either gait over the course of speed testing. This follows a pattern we have observed, whereby moderate ``damage'' to the HSA actuators can actually improve robot speed and agility (usually at the cost of some increased drift). 

One issue revealed during the tracing process is the impact of the tether on robot performance. Gait training was conducted near the center of the working area, and when the robot started to reach the edges of the enclosure, we found that the tether introduced substantial rotational and translational drift. Since the tracing was performed under closed loop control, the robot was able to compensate. However, turning speed was reduced, and switching to an untethered solution is a priority for future work.

\section{Conclusion}

In this work, we present a method that rapidly discovers mobility strategies for soft quadrupedal platforms, with minimal computation and no hand-tuning. This is accomplished through a simple yet flexible gait structure, and a tree-search based workflow that allows new gaits to be found in as little as 4 minutes. We then demonstrate how the resulting gaits can be used to trace figures and provide a closed loop control strategy which enables the quadruped to follow similar commands to a conventional differential drive robot. This expands the utility of a platform, since navigation tools for diff-drive robots are already very well developed. We also show that the gaits produced are $150\%$ faster than pre-existing hand-crafted gaits for the same robot. 

In future work, we plan to investigate automatic primitive discovery to reduce the method's reliance on human domain knowledge as well as search re-ordering as a strategy for better transfer learning. We also plan to investigate methods for detecting and adapting to shifting robot dynamics on an ongoing basis. Further development of the platform to leverage gait generation and closed loop control should enable it to perform navigation and mapping tasks using pre-existing tools from rigid robotics. 

\section*{Acknowledgements}
J.K., S.S., and T.M. acknowledge support from the Army Research Office (ARO, Grant No. W911NF-22-1-0286). P.K. and R.L.T acknowledge support from the Office of Naval Research (ONR, Grant No. N00014-22-1-2447).

\balance
\bibliographystyle{IEEEtran}
\bibliography{references}

\end{document}